# Differential Evolution with Better and Nearest Option for Function Optimization

Haozhen Dong
State Key Lab of Digital Manufacturing Equipment & Technology
Huazhong University of Science and Technology
Wuhan, China
1455921@qq.com

Liang Gao
State Key Lab of Digital Manufacturing Equipment & Technology
Huazhong University of Science and Technology
Wuhan, China
gaoliang@mail.hust.edu.cn

Xinyu Li
State Key Lab of Digital Manufacturing Equipment & Technology
Huazhong University of Science and Technology
Wuhan, China
lixinyu@mail.hust.edu.cn
Corresponding Author

Haorang Zhong
State Key Lab of Digital Manufacturing Equipment & Technology
Huazhong University of Science and Technology
Wuhan, China
1520098925@qq.com

Bing Zeng
XEMC Windpower Company Limited
Xiangtan, China
zengbing2016@126.com

*Abstract*—Differential evolution(DE) is a conventional algorithm with fast convergence speed. However, DE may be trapped in local optimal solution easily. Many researchers devote themselves to improving DE. In our previously work, whale swarm algorithm have shown its strong searching performance due to its niching based mutation strategy. Based on this fact, we propose a new DE algorithm called DE with Better and Nearest option (NbDE). In order to evaluate the performance of NbDE, NbDE is compared with several meta-heuristic algorithms on nine classical benchmark test functions with different dimensions. The results show that NbDE outperforms other algorithms in convergence speed and accuracy.

## I. INTRODUCTION

Meta-heuristic methods have become a powerful tool for numerical optimization problems, especially for those can hardly be solved by conventional mathematic method, such as travelling salesman problem [1], routing problem of wireless sensor networks (WSN) [2], etc. Normally, optimization problems in engineering often come with a given mathematical model which is featured with strong nonlinearity and multi-coupling [3], and classical mathematical method, such as Gradient method, Gauss Newton method, are gradient-based, which means that they may be trapped in local optimal solution easily. And for some problems such as multi-objective coupling problems and discrete problems, the gradient can hardly be calculated. Therefore, meta-heuristic method, such as the famous genetic algorithm (GA), differential evolution (DE), particle swarm optimization (PSO), has become a new and effective choice for solving engineering problems, for the reason that meta-heuristic method is not gradient-based and is easily to implement. Among the algorithms mentioned above, DE is the algorithm with fastest convergence speed, but it may be trapped in local optimal solution easily. In this paper, inspired by niching strategy of WSA, we propose a new DE algorithm for function optimization called differential evolution with Better and Nearest option (NbDE), which is based on the basic differential evolution (DE) and whale swarm algorithm (WSA). Here, a brief overview of DE and WSA is presented.

The famous differential evolution (DE) algorithm is proposed by Storn and Price [4], and it's designed for function optimization. Similar to the famous genetic algorithm (GA), DE also consists of three parts. Firstly, a reference vector for each individual, which can be called target vector, is created by using DE mutation strategy. Then, a crossover operation will be implemented between the target vector and the original one, and the candidate vector for each individual is generated by selecting elements from these two vectors by using crossover method. Finally, a comparison between the fitness value of the candidate vector and the original one will determine which one will transmit into the next generation. Since put forward, DE has been accepted by researchers from different fields, many researchers and engineers have proposed various ideas for using DE to solve real-world optimization problems [5][6][7]. Although DE is featured with fast convergence speed and simple mutation strategy, it often falls into local optimal solution, so improvement for DE has become a hot topic. In literature [8], Das have pointed out that recent researches of improved DE focus on strategy selection, parameter adaption, and new strategies about initialization, mutation, crossover, both of them have achieved significant results.

WSA [9][10] is a new meta-heuristic algorithm proposed by us previously, which is inspired by communicating behavior of whales. WSA uses a special mutation strategy as follows:

$$v_{i,j}^G = x_{i,j}^G + rand(0, \rho_0 e^{-\eta d_{X,Y}}) * (y_{i,j}^G - x_{i,j}^G) \qquad (1)$$

Where, $v_{i,j}^G$ denotes the j-th elements of candidate vector $v_i^G$ at G iteration and corresponding $x_{i,j}^G$ denotes the j-th elements of $x_i^G$'s position, $y_{i,j}^G$ represents the j-th element of "the Better and Nearest whale of $x_i^G$" $y_i^G$'s position at G iteration. The $rand(0, \rho_0 e^{-\eta d_{X,Y}})$ denotes creating a mutation parameter with the range from 0 to $\rho_0 e^{-\eta d_{X,Y}}$. $\rho_0$ is the ultrasound intensity that each "whale" send out, in most cases, it can be set to 2. $e$ denotes the natural constant and $\eta$ represents the attenuation coefficient. Besides, $d_{X,Y}$ denotes the Euclidean distance between **X** and **Y**. In reference [10], Zeng has proved that $\eta$ could to be set to 0 for most cases. So the Eq.1 can be simplified to the following form.

$$v_{i,j}^G = x_{i,j}^G + 2 * rand(0,2) * (y_{i,j}^G - x_{i,j}^G) \quad (2)$$

According to Eq.1 and Eq.2, we notice that a "whale" will search following its "Better and Nearest whale" which is similar to cluster method, and this can be treated as a new niching method. Simulation results have shown that WSA outperforms several classical niching methods especially in multimodal function optimization. Despite the fact that WSA can maintain population diversity in exploration process and has strong local exploration ability, drawback also exists. The convergence speed of WSA is slower than many algorithms especially in solving high-dimensions objective functions. One reason which cause this problem above is that the "Better and Nearest" option helps us shrink the range of exploration, which improves the local exploration ability but reduces convergence speed of WSA. Considering that the characteristics of DE and WSA can be complementary, we propose an idea of combining them.

The remainder of this paper can be summarized as follows: Firstly, we present the framework of NbDE. Secondly, benchmark test for comparison and configurations of comparison algorithms are introduced. And then the simulation results and their analysis are shown. The end of this paper is the conclusions of NbDE and the expectation for further research.

## II. THE FRAMEWORK OF PROPOSED NBDE

In this section, the framework of NbDE is introduced. This algorithm implements the mutation strategy "DE/rand-to-nearest & better/2", which derives from the classical DE algorithm and is inspired by WSA.

### A. Mutation

The mutation strategy "DE/rand/1" is designed for the classical DE [11][12], which is efficient for many engineer problems as we can see in literature [13]. However, drawbacks such as premature convergence also exist, which limited the further application of DE. For this reason, many researchers have proposed their solutions. R. Gamperle [14] has found that DE with "DE/best/2" strategy may outperform the original DE in many problems, and [15] proposed the mutation strategy "DE/best/1" to solve technical problems. The famous "JADE" proposed by Jingqiao Zhang [16] used "DE/current-to-pbest" combining with some other strategies, JADE has achieved a set of satisfactory results for benchmark functions.

The mutation strategy is utilized for creating the candidate vector $v_{i,G}$, "DE/rand/1" is a classical strategy which has been applied in classical DE and some other derived algorithms, and it can be described by the following function:

$$v_i^G = x_{r1}^G + F \times (x_{r2}^G - x_{r3}^G), r1 \neq r2 \neq r3 \neq i \quad (3)$$

Where G denotes the number of iteration, r1, r2, r3 represent the individuals ID and they are selected with randomly strategy. F denotes mutation operator parameter which is used for scaling the differential vector.

As mentioned above, the WSA mutation strategy can be summarized as Eq.2. Based on mutation strategies of DE and WSA, we proposed a hybrid mutation strategy as follows:

$$v_i^G = x_{r1}^G + rand(0,1) \times (x_{r2}^G - x_i^G) + rand(0,1) \times (y_i^G - x_i^G) \quad (4)$$

Similar to mutation strategies above, G is the number of current iteration, r1, r2 represent the random individuals ID, $y_i^G$ denotes the nearest individual with better fitness value of $x_i^G$ and when $x_i^G$ is the best individual currently we will choose a random individual for $y_i^G$.

### B. Crossover

Different from classical DE, we provided a strategy which combines binary crossover, exponential crossover and non-crossover operator. Firstly, a random number is generated to determine which crossover operator will be selected. Then the crossover operation is implemented between $v_{i,G}$ and $x_{i,G}$, and the final candidate $v_{i,G}$ will be generated.

*1) Binary crossover*

The binary crossover of NbDE can be described as follows:

$$v_{i,j}^G = \begin{cases} v_{i,j}^G & if\ rand(0,1) \leq CR \\ x_{i,j}^G & otherwise \end{cases}, i = 1,2 \cdots NP; j = 1,2 \cdots D \quad (5)$$

Where $rand(0,1)$ denotes a random number with the range from 0 to 1, NP is the population size, CR represents the crossover control parameter and D denotes the dimension of decision variable.

TABLE.I THE PSEUDO CODE OF NBDE

| Input: Objective function to be solved, options of the NbDE. Output: An optimal solution. |
|---|
| 1: Begin & Initialization |
| 2: Initialize a group of individuals; |
| 3: Evaluate each individual; |
| 4: Judge whether termination criterion is satisfied |
| 5: For i=1 to NP |
| 6: Create a new individual $v_i^G$ by Eq.4; |
| 7: Crossover $v_i^G$ with $x_i^G$; |
| 8: Evaluate the new individual $v_i^G$; |
| 9: If $f(v_i^G) < f(x_i^G)$ |
| 10: $x_i^{G+1} = v_i^G$; |
| 11: End If |
| 12: End for |
| 14: End |

*2) Exponential crossover*

The exponential crossover of NbDE can be expressed as follows:

$$v_{i,j}^G = \begin{cases} v_{i,j}^G & if\ j_{down} \leq j \leq j_{up} \\ x_{i,j}^G & otherwise \end{cases}, j_{up} = j_{down} + randi(D) \quad (6)$$

$$randi(D) = sum(rand(1,D) \leq CR) \quad (7)$$

Where $j_{up}$ and $j_{down}$ denote the start and end dimension of exponential crossover, $randi(D)$ represents the number of elements no more than CR in random vector $rand(1,D)$. When $j_{up} > D$, this operator can be written in this form:

$$v_{i,j}^G = \begin{cases} v_{i,j}^G & if j_{down} \leq j \text{ or } j \leq j_{up} - D \\ x_{i,j}^G & otherwise \end{cases}, i = ,2 \cdots NP; j = 1,2 \cdots D \quad (8)$$

It must be noticed that, in this algorithm CR is a random number with the range 0.4-0.9 instead of a constant value.

*C. Selection*

When a candidate individual $v_i^G$ is created by previous steps, there will be a comparison between $f(v_i^G)$ and $f(x_i^G)$, the individual with better fitness value will "survive". This greedy selection strategy can be shown as follows:

$$x_i^{G+1} = \begin{cases} v_i^G & if\ f(v_i^G) \leq f(x_i^G) \\ x_i^G & otherwise \end{cases}, i = 1,2 \cdots NP \quad (9)$$

## III. SIMULATION OF NBDE

NbDE is tested with a set of 9 benchmark functions with different dimensions (D=10, D=30, D=50) as shown in Table. 2, test functions F1-F5 are unimodal benchmark functions and F6-F9 are multimodal functions, based on the results of these benchmark functions, we can get a general evaluation of NbDE's performance. What's more, NbDE is compared with other 5 algorithms, which include the famous adaptive DE algorithm JADE, the classic DE/rand/1, the DE/best/2, the classic genetic algorithm (GA), the classic particle swarm optimization (PSO) and WSA. For fair comparison, all methods are allowed to evaluate the objection functions with maximum 10000D times. Based on the suggestions from original papers, other configurations of all the algorithms mentioned above are shown as follows:

- NbDE: NP=40;
- DE/rand/1: F=0.5; CR=0.9; NP=30(D=10); NP=100(D=30); NP=200(D=50).
- DE/best/2: F=0.5; CR=0.9; NP=30(D=10); NP=100(D=30); NP=200(D=50).
- JADE [16]: p=0.05; c=0.1; CR=0.9; NP=30(D=10); NP=100(D=30); NP=200(D=50).
- WSA [10]: $\eta = 0$; NP=40.
- GA [17]: CP=0.95; MP=0.05; NP=40(D=10); NP=100(D=30); NP=200(D=50).
- PSO [18]: C1=2.05; C2=2.05; vMax=2; vMin=-2; K=0.729; NP=40(D=10); NP=100(D=30); NP=200(D=50).

In these tests, the only parameter we should set for NbDE benchmark test is the number of population, for most cases without too much decision variables, we can use the same NP. All of these methods are implemented with Matlab 2014b and executed on a personal PC with 3.4 GHz Intel Xeon E3-1230-V5 processor, 16 GB RAM.

All of the test functions mentioned above are simulated by each algorithm with 50 independent runs, and the statistical results shown in Table. 3-8 which are organized by the dimensions and evaluation indexes of test functions. In these tables, four important statistical indexes included the mean value, standard deviation (STD) of the results, the success rate (SR) and its rank are given. For success rate statistics, the success value was set to 1E-02 for F5, while 1E-08 for others. Computing time is not given in this comparison for the reason that it is not a criterion to be investigated here. NbDE is expected to be a little slower than the classical DE because the searching of the "Better and Nearest" individual in each evaluation will cost a little time.

TABLE. II BENCHMARK FUNCTION DEFINITIONS

| Fn | Test Function Name | Bounds | Optimum value |
|---|---|---|---|
| F1 | Zakharov | $[-100,100]^D$ | 0 |
| F2 | Schwefel 2.22 | $[-10,10]^D$ | 0 |
| F3 | Schwefel 2.21 | $[-100,100]^D$ | 0 |
| F4 | Rosenbrock | $[-30,30]^D$ | 0 |
| F5 | Noise Quartic | $[-1.28,1.28]^D$ | 0 |
| F6 | Schwefel 2.26 | $[-500,500]^D$ | -418.9828872724339D |
| F7 | Rastrigin | $[-5.12,5.12]^D$ | 0 |
| F8 | Ackley | $[-32,32]^D$ | 0 |
| F9 | Griewank | $[-600,600]^D$ | 0 |

*A. Success Rate*

As we can see in Table.3-6, the success rates and ranks of NbDE and other methods on benchmark functions are shown. When two algorithms get the same success rate on a benchmark function, they will get the same rank score. The last row of these tables shows the sum of each algorithm's rank score, which can represent overall performance of different methods on benchmark functions.

TABLE.III SR AND RANKS OF ALL ALGORITHMS WHEN D=10

| Fun | NbDE | DE/rand/1 | DE/best/2 | JADE | WSA | GA | PSO |
|---|---|---|---|---|---|---|---|
| F1 | **1/1** | 0/5 | 0/5 | 0.26/4 | 0.98/3 | 0/5 | 1/1 |
| F2 | **1/1** | **1/1** | **1/1** | **1/1** | 0.86/5 | 0/6 | 0/6 |
| F3 | **1/1** | 0/4 | **1/1** | 0.02/3 | 0/4 | 0/4 | 0/4 |
| F4 | **0.98/1** | 0/3 | 0/3 | 0/3 | 0/3 | 0/3 | 0.74/2 |
| F5 | **1/1** | **1/1** | **1/1** | **1/1** | 0.96/6 | 0.92/7 | **1/1** |
| F6 | **1/1** | 0.46/2 | 0.42/3 | 0.24/4 | 0/5 | 0/5 | 0/5 |
| F7 | **1/1** | 0.14/4 | 0.98/2 | 0.94/3 | 0/5 | 0/5 | 0/5 |
| F8 | **1/1** | **1/1** | **1/1** | 0.9/4 | 0.72/5 | 0/7 | 0.52/6 |
| F9 | 0.68/2 | 0.56/3 | **0.9/1** | 0.46/4 | 0/5 | 0/5 | 0/5 |
| Rank Score | **10** | 24 | 18 | 27 | 41 | 47 | 35 |

TABLE.IV SR AND RANKS OF ALL ALGORITHMS WHEN D=30

| Fun | NbDE | DE/rand/1 | DE/best/2 | JADE | WSA | GA | PSO |
|---|---|---|---|---|---|---|---|
| F1 | **1/1** | 0/2 | 0/2 | 0/2 | 0/2 | 0/2 | 0/2 |
| F2 | **1/1** | **1/1** | **1/1** | **1/1** | 0.5/5 | 0/6 | 0/6 |
| F3 | **1/1** | 0/2 | 0/2 | 0/2 | 0/2 | 0/2 | 0/2 |
| F4 | **0.94/1** | 0/2 | 0/2 | 0/2 | 0/2 | 0/2 | 0/2 |
| F5 | **1/1** | 0/7 | **1/1** | **1/1** | 0.2/6 | 0.66/5 | 0.98/4 |
| F6 | **1/1** | 0.54/3 | 0.58/2 | 0/4 | 0/4 | 0/4 | 0/4 |
| F7 | 0.92/2 | 0.18/3 | **1/1** | 0/4 | 0/4 | 0/4 | 0/4 |
| F8 | **1/1** | **1/1** | **1/1** | 0.1/4 | 0/5 | 0/5 | 0/5 |
| F9 | 0.9/3 | 1/1 | **1/1** | 0.84/4 | 0.44/5 | 0/6 | 0/6 |
| Rank Score | **14** | 22 | 14 | 24 | 35 | 36 | 34 |

As we can see in Table.3-6, NbDE get the 100% success rate on F1, F2, F3, F5, F6, F7, F8 when D=10, on F1, F2, F3, F5, F6, F8 when D=30 and on F1, F2, F5, F8 when D=50. For most functions, NbDE get the highest success rate, but we notice that the success rate of NbDE on F9 is slightly lower than those of DE/best/2 when D=10. D=30 and D=50 and DE when D=30, and the success rate on F7 is a little bit lower than that of DE/best/2 when D=30. Besides, NbDE performs much better than other algorithms.

TABLE.V SR AND RANKS OF ALL ALGORITHMS WHEN D=50

| Fun | NbDE | DE/rand/1 | DE/best/2 | JADE | WSA | GA | PSO |
|---|---|---|---|---|---|---|---|
| F1 | **1/1** | 0/2 | 0/2 | 0/2 | 0/2 | 0/2 | 0/2 |
| F2 | **1/1** | 0/5 | **1/1** | 0.38/3 | 0.14/4 | 0/5 | 0/5 |
| F3 | **0/1** | **0/1** | **0/1** | **0/1** | **0/1** | **0/1** | **0/1** |
| F4 | **0.3/1** | 0/2 | 0/2 | 0/2 | 0/2 | 0/2 | 0/2 |
| F5 | **1/1** | 0/6 | 0.02/5 | **1/1** | 0/6 | 0.14/4 | 0.92/3 |
| F6 | **0.68/1** | 0.28/2 | 0.02/3 | 0/4 | 0/4 | 0/4 | 0/4 |
| F7 | **0/1** | **0/1** | **0/1** | **0/1** | **0/1** | **0/1** | **0/1** |
| F8 | **1/1** | 0/3 | 1/1 | 0/3 | 0/3 | 0/3 | 0/3 |
| F9 | 0.96/2 | 0/5 | **1/1** | 0.14/4 | 0.32/3 | 0/5 | 0/5 |
| Rank Score | **10** | 27 | 17 | 21 | 26 | 27 | 26 |

From Table.3 and Table.5, we can conclude that NbDE performs best on success rate of benchmark function test when D=10 and D=50, because the rank score of NbDE is much better than those of other methods. We have also noticed that NbDE and DE/best/2 got the same rank score when D=30. But we can see in Table.4 the rank of DE/best/2 on F1, F3, F4 is 2 while the SR of DE/best/2 is 0 which means that DE/best/2 cannot get exactly results on these benchmark functions, but NbDE can get exactly results in most cases, so we can infer that NbDE outperform DE/best/2 when D=30. Therefore, we can conclude that NbDE outperforms other methods on index of success rate in benchmark functions test.

B. *Quality of Optima Found*

In this part, NbDE is compared with other methods in the index of the optima found accuracy. As we can see from Table. 6-8, we have found that NbDE gets the optima found with best accuracy on F1, F3, F4, F6, F7, F8, F9 when D=10, on F1, F2, F3, F5, F6, F8 when D=30, on F1, F2, F3, F4, F6, F8 when D=50.

What more, We have also noticed that the performance index of F9 reached by NbDE is relatively small(less than 5E-3), considering that the success (SR) of F9 when D=30 and D=50 are even higher than those when D=10, so we can conclude that NbDE has jumped out of local optimal and needs more iterations for convergence in these cases. Based on this analysis, we can also make the similar conclusion for F4 when D=50. Table.7 and Table.8 have shown that DE/best/2 get the best result for F7 and F9 when D=30 and D=50, but it can hardly get a feasible solution for F1, F3, F4 when D=30, F1, F3, F4, F5, F6, F7 when D=50. So we can also conclude that NbDE outperforms DE/best/2. We can also find that JADE underperforms NbDE in most cases while it can get satisfactory solutions when the number of iterations is enough[16], so we can conclude that JADE converge slower than NbDE. GA and PSO is classical meta-heuristic algorithms for function optimization, in this experiment we can observe that NbDE outperforms GA and PSO significantly. WSA is the new meta-heuristic method we proposed previously, it's featured with strong local exploration ability but its convergence speed is relatively slow especially for high dimensions problems which can also be demonstrated in this simulation. From inferences mentioned above, we can find that NbDE outperforms other algorithms and is featured with fast convergence speed and small population size because of the improved mutation and crossover strategy.

TABLE.VI QUALITY OF OPTIMA FOUND OF ALL ALGORITHMS WHEN D=10 (MEASUREMENT: MEAN/STD)

| Fun | NbDE | DE/rand/1 | DE/best/2 | JADE | WSA | GA | PSO |
|---|---|---|---|---|---|---|---|
| F1 | **3.3E-40/1.6E-39** | 6.8E+3/2.8E+3 | 2.7E+2/1.4E+2 | 1.5E-1/8.2E-1 | 5.3E-10/3.0E-9 | 2.1E+4/7.7E+3 | 4.8E-25/2.7E-24 |
| F2 | 5.3E-39/7.5E-39 | 5.2E-40/4.9E-40 | **7.8E-47/6.5E-47** | 3.2E-12/2.2E-11 | 2.26E+0/1.1E-2 | 1.3E-3/7.3E-4 | 1.4E-2/1.5E-2 |
| F3 | **6.6E-23/1.1E-22** | 3.9E-06/3.0E-6 | 1.2E-12/6.4E-13 | 2.6E-2/4.5E-3 | 4.0E+0/2.2E+0 | 4.1E-1/2.8E-1 | 1.8E-2/1.5E-2 |
| F4 | **8.0E-2/5.6E-1** | 6.7E+0/8.3E+0 | 1.0E+0/1.9E+0 | 5.9E+0/2.3E+0 | 2.0E+1/3.2E+1 | 1.4E+1/8.8E+0 | 1.0E+0/1.8E+0 |
| F5 | 1.2E-3/5.2E-4 | 3.2E-3/1.2E-3 | 1.4E-3/5.2E-4 | **3.8E-4/2.5E-4** | 4.1E-3/3.0E-3 | 5.6E-3/2.6E-3 | 1.9E-3/1.6E-3 |
| F6 | **-418.98E+1/1.8E-12** | -409.85E+1/1.2E+2 | -409.93E+1/9.1E+1 | -402.88E+1/1.3E+2 | -336.3E+1/2.7E+2 | -134.86E+1/2.5E+2 | -221.21E+1/5.2E+2 |
| F7 | **0/0** | 1.4E+0/1.2E+0 | 2.0E-2/1.4E-1 | 6.0E-2/2.3E-1 | 9.4E+0/5.8E+0 | 7.3E-5/1.4E-4 | 1.0E+1/3.6E+0 |
| F8 | **4.2E-15/8.5E-16** | 4.5E-15/5.0E-16 | 4.4E-15/0 | 2.3E-2/1.6E-1 | 3.7E-1/6.8E-1 | 5.9E-3/3.3E-3 | 7.9E-1/9.3E-1 |
| F9 | **2.1E-4/1.1E-3** | 3.3E-3/5.3E-3 | 7.9E-4/2.4E-3 | 9.5E-3/1.1E-2 | 8.6E-2/6.7E-2 | 6.6E-2/2.6E-2 | 4.5E-1/2.7E-1 |

TABLE.VII QUALITY OF OPTIMA FOUND OF ALL ALGORITHMS WHEN D=30 (MEASUREMENT: MEAN/STD)

| Fun | NbDE | DE/rand/1 | DE/best/2 | JADE | WSA | GA | PSO |
|---|---|---|---|---|---|---|---|
| F1 | **5.4E-33/8.6E-33** | 6.3E+4/7.5E+3 | 1.6E+4/2.8E+3 | 4.0E-2/6.2E-2 | 2.18E+2/1.5E+3 | 8.1E+4/1.6E+4 | 4.6E-3/3.0E-3 |
| F2 | **1.1E-50/1.6E-50** | 3.1E-15/7.23E-16 | 2.3E-27/5.7E-28 | 4.7E-11/3.3E-10 | 2.39E+0/4.7E+0 | 2.2E-2/8.6E-3 | 6.3E-1/5.6E-1 |
| F3 | **3.9E-17/8.8E-17** | 5.5E+0/4.8E-1 | 5.6E-4/9.0E-5 | 8.0E-1/4.1E-1 | 2.5E+1/6.1E+0 | 6.1E+0/1.5E+1 | 2.5E-1/2.1E-1 |
| F4 | **2.3E-2/9.5E-1** | 6.4E+1/2.0E+1 | 2.7E+1/1.3E+1 | 3.6E+1/1.8E+1 | 1.9E+3/1.3E+4 | 2.5E+2/2.3E+2 | 2.8E+1/1.0E+1 |
| F5 | **2.0E-3/6.4E-4** | 2.7E-2/5.5E-3 | 6.1E-3/1.4E-3 | 8.3E-4/2.7E-4 | 1.1E-3/3.7E-1 | 9.2E-3/3E-3 | 4.0E-3/2.4E-3 |
| F6 | **-125.69E+2/7.3E-12** | 124.89E+2/1.2E+2 | 125.08E+2/1.0E+2 | -107.87E+2/7.0E+2 | -729.15E+1/7.0E+2 | -261.42E+1/4.0E+2 | -628.17E+1/1.2E+3 |
| F7 | 1.0E-1/3.6E-1 | 7.8E-1/8.5E-1 | **0/0** | 4.5E+1/5.0E+0 | 1.1E+2/3.5E+1 | 1.3E-1/3.2E-1 | 4.4E+1/1.4E+1 |
| F8 | **6.2E-15/1.8E-15** | 5.0E-10/4.5E-10 | 8.9E-15/1.6E-15 | 1.4E-5/5.7E-5 | 5.6E+0/2.2E+0 | 3.4E-2/1.1E-2 | 2.6E+0/8.0E-1 |
| F9 | 1.0E-3/3.3E-3 | 0/0 | **0/0** | 7.9E-4/3.2E-3 | 1.5E-2/1.9E-2 | 1.1E-7/7.7E-2 | 1.1E-2/1.3E-2 |

TABLE.VIII QUALITY OF OPTIMA FOUND OF ALL ALGORITHMS WHEN D=50 (MEASUREMENT: MEAN/STD)

| Fun | NbDE | DE/rand/1 | DE/best/2 | JADE | WSA | GA | PSO |
|---|---|---|---|---|---|---|---|
| F1 | **1.9E-21/7.3E-21** | 1.3E+5/1.2E+4 | 6.1E+4/5.3E+3 | 8.1E+3/1.4E+3 | 4.1E+3/4.1E+3 | 1.4E+5/2.4E+4 | 4.65E-2/1.5E-2 |
| F2 | **1.4E-61/1.6E-61** | 2.7E-6/2.8E-7 | 2.7E-17/3.6E-18 | 3.2E-6/1.2E-5 | 1.5E+1/2.6E+1 | 2.0E-5/5.1E-2 | 1.6E+0/8.5E-1 |
| F3 | **8.9E-2/1.7E-1** | 4.2E+1/1.6E+0 | 1.4E-1/1.4E-2 | 1.6E+0/4.7E-1 | 3.6E+1/7.2E+0 | 2.4E+1/5.3E+0 | 6.5E-1/3.0E-1 |
| F4 | **2.4E-1/9.6E-1** | 2.4E+2/1.8E+1 | 4.7E+1/1.1E+1 | 7.0E+1/2.8E+1 | 3.7E+3/1.8E+4 | 7.8E+2/3.6E+2 | 6.1E+1/2.9E+1 |
| F5 | 3.5E-3/1.0E-3 | 1.3E-1/1.6E-1 | 1.3E-2/2.0E-3 | **1.3E-3/3.1E-4** | 1.2E+0/6.8E+0 | 1.4E-2/3.8E-3 | 5.1E-3/3.7E-3 |
| F6 | **-209.04E+2/7.1E+1** | -207.70E+2/2.1E+2 | -196.81E+2/4.6E+2 | -111.50E+2/4.4E+2 | -111.25E+2/1.0E+3 | -383.82E+1/6.0E+2 | -102.79E+2/1.5E+3 |
| F7 | 3.7E+0/1.7E+0 | 6.5E+2/6.5+0 | **4.5E-2/4.1E-2** | 1.7E+2/8.7E+0 | 2.6E+2/4.6E+1 | 1.3E+1/3.8E+0 | 6.5E+1/1.5E+1 |
| F8 | **9.3E-15/2.7E-15** | 9.3E-2/2.1E-2 | 2.1E-14/1.4E-15 | 2.2E-4/1.8E-4 | 1.2E+1/3.4E+0 | 8.0E-1/4.9E-1 | 3.3E+0/4.7E-1 |
| F9 | 7.9E-4/4.4E-3 | 7.1E-6/1.6E-6 | **0/0** | 9.2E-4/3.1E-3 | 4.1E-2/1.0E-1 | 9.8E-1/8.1E-2 | 7.1E-3/7.5E-3 |

## IV. Conclusion

In this paper, inspired by whale swarm algorithm and differential evolution, a meta-heuristic algorithm for function optimization called differential evolution with better and nearest option (NbDE), is proposed. NbDE is compared with several other meta-heuristic methods on success rate and optima found accuracy of benchmark functions. Simulation results have shown that NbDE outperforms other algorithms in overall index of benchmark functions and have demonstrated the effectiveness of NbDE. In the future we will pay more attention on the following aspects:

1) NbDE application in humanoid robot, such as the design of mechanical structure and the optimization of fuzzy logic controller;
2) Improvement of NbDE such as parallel computing optimization, parameter self-adjustment.

## Acknowledgment

This research was supported by the National Natural Science Foundation for Distinguished Young Scholars of China under Grant No.51825502, National Natural Science Foundation of China (NSFC) (51775216, and 51721092), Natural Science Foundation of Hubei Province (2018CFA078) and the Program for HUST Academic Frontier Youth Team.

## References

[1] Mahi, M., Baykan, Ö.K., Kodaz, H.: A new hybrid method based on particle swarm optimization, ant colony optimization and 3-opt algorithms for traveling salesman problem. Appl. Soft Comput. 30, 484–490 (2015)

[2] Zeng, B., Dong, Y.: An improved harmony search based energy-efficient routing algorithm for wireless sensor networks. Appl. Soft Comput. 41, 135–147 (2016)

[3] Strogatz, S. (2015). Nonlinear Dynamics and Chaos. Boca Raton: CRC Press.Storn, R., Price, K.: Differential Evolution - A Simple and Efficient Adaptive Scheme for Global Optimization Over Continuous Spaces. ICSI, Berkeley (1995)

[5] Qing, A.: Dynamic differential evolution strategy and applications in electromagnetic inverse scattering problems. IEEE Trans. Geosci. Remote Sens. 44(1), 116–125 (2006)

[6] Gao, Z., Pan, Z., Gao, J.: A new highly efficient differential evolution scheme and its application to waveform inversion. IEEE Geosci. Remote Sens. Lett. 11(10), 1702–1706 (2014)

[7] Das, S., Suganthan, P.N.: Differential evolution: a survey of the state-of-the-art. IEEE Trans. Evol. Comput. 15(1), 4–31 (2011)

[8] Das, Swagatam, Sankha Subhra Mullick, and Ponnuthurai N. Suganthan. "Recent advances in differential evolution–an updated survey." Swarm and Evolutionary Computation 27 (2016): 1-30.

[9] Zeng, Bing, L. Gao, and X. Li. Whale Swarm Algorithm for Function Optimization. Intelligent Computing Theories and Application. 2017.

[10] Zeng, Bing, et al. "Whale swarm algorithm with the mechanism of identifying and escaping from extreme points for multimodal function optimization." Neural Computing and Applications: 1-21.

[11] R. Storn and K. Price, "Differential evolution a simple and efficient heuristic for global optimization over continuous spaces," J. Global Optimization, vol. 11, no. 4, pp. 341–359, 1997.

[12] K. V. Price, R. M. Storn, and J. A. Lampinen, Differential Evolution: A Practical Approach to Global Optimization, 1st ed. New York: SpringerVerlag, Dec. 2005.

[13] "Differential evolution for multiobjective optimization," in Proc. IEEE Congr. Evol. Comput., Dec. 2003, pp. 2696–2703.

[14] R. Gamperle, S. D. Muller, and P. Koumoutsakos, "A parameter study for differential evolution," in Proc. Advances Intell. Syst., Fuzzy Syst., Evol. Comput., Crete, Greece, 2002, pp. 293–298.

[15] U. Pahner and K. Hameyer, "Adaptive coupling of differential evolution and multiquadrics approximation for the tuning of the optimization process," IEEE Trans. Magnetics, vol. 36, no. 4, pp. 1047–1051, Jul. 2000.

[16] J., Z. and C.S. A., JADE: Adaptive Differential Evolution With Optional External Archive. IEEE Transactions on Evolutionary Computation, 2009. 13(5): p. 945-958.

[17] Yuan, X., et al., A Genetic Algorithm-Based, Dynamic Clustering Method Towards Improved WSN Longevity. Journal of Network and Systems Management, 2017. 25(1): p. 21-46.

[18] Bansal J.C. (2019) Particle Swarm Optimization. In: Bansal J., Singh P., Pal N. (eds) Evolutionary and Swarm Intelligence Algorithms. Studies in Computational Intelligence, vol 779. Springer, Cham